\documentclass[runningheads]{llncs}
\usepackage{url}
\usepackage[T1]{fontenc}
\usepackage{graphicx}
\usepackage{bbding}
\begin{document}
\title{Efficient Reasoning Distillation: Small Video-Language Models via Synthetic CoT and Difficulty-Aware Fine-Tuning}
\titlerunning{Efficient Reasoning Distillation for Small VLMs}
%
\author{Mantek Singh\inst{1}\Envelope \orcidID{0009-0008-1556-7452} \and
Jeshwanth Challagundla\inst{3}\orcidID{0009-0005-5106-3743}\and
Siddharth Raina\inst{4}\orcidID{0009-0001-8745-0199} \and
Jasmin Jarsania\inst{2}\orcidID{0009-0005-4713-3614}}
\authorrunning{M. Singh et al.}
\institute{Liverpool John Moores University, Liverpool, England \\
\email{mantek.singh2@gmail.com} \and
University of Texas at Arlington, Texas, USA \and
Carnegie Mellon University, Pittsburgh, USA \and 
Meta, Sunnyvale, USA \\
}
\maketitle              

\begin{abstract}
We present an efficient method to distill reasoning capabilities into compact video-language models (VLMs) for video question answering (VideoQA). Our approach fine-tunes a 2B-parameter model using only $\sim$900 uncertainty-selected examples, each augmented with synthetic chain-of-thought (CoT) rationales generated by a 4B teacher. Despite its minimal compute cost---under two hours on a single A100 GPU---our method enables the 2B model to outperform VLMs up to 4$\times$ larger, and generalize across CinePile, ActivityNet-QA, and MLVU, approaching the performance of its own 4B teacher.

A key finding is that placing CoT rationales \emph{after} the answer---contrary to standard prompting---substantially improves reasoning in compact models. This insight challenges prevailing CoT conventions and reveals new alignment strategies under limited model capacity. Our findings offer a practical blueprint for training deployable, reasoning-rich VLMs suited for mobile and edge applications.
\end{abstract}
\section{Introduction}
\label{sec:intro}

The surge in video content requires efficient video-language models (VLMs) for edge devices, yet current models are prohibitively large for deployment in resource-constrained environments~\cite{sharshar2025visionlanguagemodelsedgenetworks,gong2025surveyvideoanalyticscloudedgeterminal}. This paper addresses the need for compact VLMs with strong reasoning capabilities for video question answering (VideoQA). Traditional video QA datasets are expensive, privacy-sensitive, and often biased~\cite{schmarje2022annotationenoughdatacentricimage,zhang2023deeplongtailedlearningsurvey,Geirhos_2020}, making synthetic data a scalable, privacy-preserving alternative for data augmentation~\cite{Mumuni_2024,zhang2024videoinstructiontuningsynthetic}.

We propose a sample-efficient framework to distill reasoning from a 4B teacher (InternVL2-4B) into a 2B student model (InternVL2-2B). Our approach combines two core strategies:
\begin{itemize}
    \item \textbf{Synthetic Chain-of-Thought (CoT) Generation}: Inspired by LLMs~\cite{wei2023chainofthoughtpromptingelicitsreasoning,zhang2024multimodalchainofthoughtreasoninglanguage,chen2024measuringimprovingchainofthoughtreasoning}, we use the 4B VLM to generate concise CoT rationales. This obviates the need for human annotation~\cite{li-etal-2023-symbolic,zelikman2022starbootstrappingreasoningreasoning,li2025smallmodelsstrugglelearn} and allows us to explore formats suitable for compact models.

    \item \textbf{Hard Sample Selection for Fine-Tuning}: Instead of using large datasets, we target high-value training points identified via model-estimated difficulty and top-2 prediction margin (uncertainty-based sampling)~\cite{settles.tr09,Niekerk_2024}, aligning with hard example mining~\cite{shrivastava2016trainingregionbasedobjectdetectors,9392296}.
\end{itemize}

We validate our method on the CinePile benchmark~\cite{rawal2024cinepilelongvideoquestion} and test generalization on ActivityNet-QA~\cite{yu2019activitynetqadatasetunderstandingcomplex} and MLVU~\cite{zhou2025mlvu}. Our key findings are:
\begin{itemize}
    \item Synthetic CoT rationales, even from a small 4B teacher, significantly improve performance.
    \item Top-2 margin sampling consistently outperforms other selection strategies.
    \item Placing rationales \emph{after} the answer substantially boosts reasoning in small models, contrary to standard CoT prompting.
    \item Fine-tuning on only 900 examples (<0.5\% of training data) yields large, efficient gains.
    \item The improvements generalize strongly across unseen datasets and domains.
\end{itemize}

To our knowledge, this is the first work to combine small-scale CoT distillation, uncertainty-guided sampling, and format-aware rationale design for compact VLMs, offering a practical recipe for reasoning-rich alignment under real-world deployment constraints.

\textit{Additional results, ablations, and qualitative examples are provided in the supplementary material.}
\section{Related Work}
\label{sec:literature_review}

\subsection{Synthetic Data and Knowledge Distillation}

Synthetic data, generated via methods like generative models~\cite{goodfellow2014generativeadversarialnetworks,kingma2022autoencodingvariationalbayes,rombach2022highresolutionimagesynthesislatent}, addresses data scarcity and privacy limitations in vision~\cite{Mumuni_2024,richter2016playingdatagroundtruth,7780721}. While recent VLMs generate question-answer pairs directly from visual inputs~\cite{liu2024improvedbaselinesvisualinstruction,zhang2024videoinstructiontuningsynthetic}, our work emphasizes distilling reasoning-focused synthetic rationales via Chain-of-Thought (CoT) rather than direct QA generation.

Knowledge distillation transfers representations from large "teacher" to smaller "student" models~\cite{hinton2015distillingknowledgeneuralnetwork,romero2015fitnetshintsdeepnets,sanh2020distilbertdistilledversionbert}. However, small models ($<$3B) often struggle to absorb complex supervision due to limited capacity~\cite{li2025smallmodelsstrugglelearn}. While methods like AoTD~\cite{AoTD} introduced large-scale video reasoning distillation, they rely on extensive supervision. We instead show that strong reasoning gains are possible with a modest 4B teacher and only 900 training examples, enabling practical edge deployment.

\subsection{Chain-of-Thought Reasoning and Hard Sample Selection}

CoT prompting improves reasoning by encouraging intermediate steps~\cite{wei2023chainofthoughtpromptingelicitsreasoning} and has been extended to multimodal VQA~\cite{zhang2024multimodalchainofthoughtreasoninglanguage,tan2024boostingpowersmallmultimodal}. Recent works distill symbolic reasoning into large VLMs~\cite{hu2024visualprogramdistillation,shi2025agentofthought}. In contrast, we focus on a low-resource regime, fine-tuning a 2B model with only 900 concise CoTs from a 4B teacher. We uniquely investigate CoT \emph{positioning} and rationale format for compact models, extending image QA CoT distillation~\cite{VPD} to long-form video with an emphasis on post-hoc rationales and format-aware alignment.

Hard sample selection accelerates convergence by focusing on uncertain samples near model decision boundaries~\cite{shrivastava2016trainingregionbasedobjectdetectors}. We apply margin-based uncertainty sampling~\cite{settles.tr09} to target ambiguous, high-value examples. Unlike typical easy-to-hard curriculum learning~\cite{bengio_curriculum_2009,wang2021surveycurriculumlearning}, we adopt a hard-first approach, which we find highly effective when paired with synthetic CoT rationales.

\subsection{VideoQA Datasets}

We evaluate our method on several benchmarks, summarized in Table~\ref{tab:vqa_datasets}. Our primary dataset for fine-tuning is the long-form, temporally rich CinePile~\cite{rawal2024cinepilelongvideoquestion}. We test zero-shot generalization on the diverse web videos of ActivityNet-QA~\cite{yu2019activitynetqadatasetunderstandingcomplex} and the compositional, multi-task design of MLVU~\cite{zhou2025mlvu}. Our focus on these datasets distinguishes our work from evaluations on benchmarks with shorter coverage like MSVD-QA~\cite{xu2017video} and TVQA~\cite{lei2019tvqalocalizedcompositionalvideo}, or those with different long-context tasks such as EgoSchema~\cite{mangalam2023egoschema} and LongVideoBench~\cite{wu2024longvideobench}.

\begin{table}[h]
\centering
\caption{Comparison of video QA datasets.}
\begin{tabular}{l|r|r|l}
\hline
Dataset & QA Pairs & Length & Focus \\
\hline
\textbf{CinePile} & $\sim$305k & $\sim$160s & Long Vid, Temporal \\
\textbf{ActNet QA} & 58k & $\sim$180s & Web Vid, Temporal \\
MSVD-QA & 50k & $\sim$10s & Short Vid, Desc. \\
TVQA & 152k & $\sim$76s & TV, Dialogue+Visual \\
\hline
\end{tabular}
\label{tab:vqa_datasets}
\end{table}
\section{Methodology}
\label{sec:methodology}

\begin{figure*}[t]
    \centering
    \includegraphics[width=\textwidth]{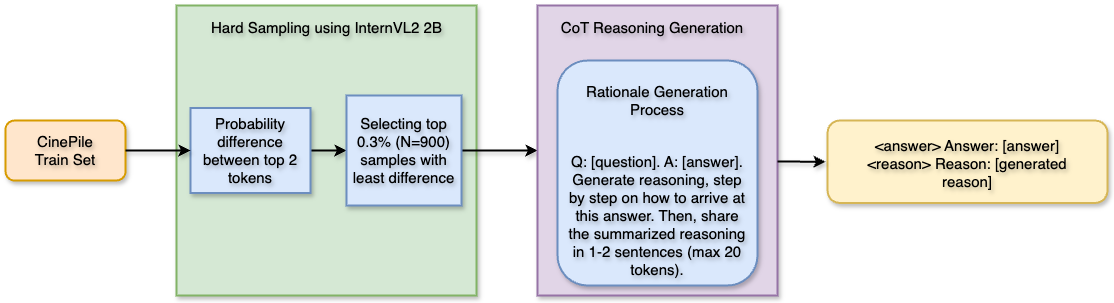}
    \caption{Overview of methodology for improving VLM using synthetic Chain-of-Thought reasoning and hard sample selection.}
    \label{fig:example}
\end{figure*}

Our goal is to enhance InternVL2-2B's video reasoning capabilities through synthetic data and lightweight fine-tuning, optimized for edge deployment. Our pipeline, shown in Fig.~\ref{fig:example}, integrates hard sample selection, synthetic CoT rationale generation, and efficient supervision.

\subsection{Hard Sample Selection and Rationale Generation}

To maximize fine-tuning impact, we first identify $N$ examples where InternVL2-2B is most uncertain. We explore three strategies: random sampling, model-estimated difficulty (ranking by negative log-probability), and top-2 margin sampling (selecting samples with the smallest gap between the top two predictions~\cite{settles.tr09,Niekerk_2024}). We found performance saturated near $N{=}700$ (see Sec.~\ref{sec:exp_results}) and chose $N{=}900$ to ensure robustness, corresponding to $<0.5\%$ of the full dataset.

For these 900 samples, we generate short, structured CoT rationales using a 4B teacher (InternVL2-4B), avoiding the verbosity of larger models~\cite{li2025smallmodelsstrugglelearn}. We use the prompt: \texttt{Q: [question]. A: [answer]. Generate reasoning...summarize it in 1--2 sentences (max 20 tokens).} Rationales are capped at 20 tokens, and manual inspection confirmed 85\% validity; we apply filters to remove malformed samples.

\subsection{Fine-Tuning and Evaluation}

We fine-tune InternVL2-2B on the $N=900$ augmented samples. We tested placing CoTs before the answer, as in standard prompting~\cite{wei2023chainofthoughtpromptingelicitsreasoning}, but observed degraded performance (e.g., final-option bias). Providing the rationale as input context ("Explanation: [Rationale]...") also yielded lower gains. In contrast, placing the rationale \emph{after} the answer led to strong gains, so we adopt the "Answer-then-Rationale" format: \texttt{<answer> [Answer]. <reason> [Rationale Text]}. We found concise, 1-2 sentence rationales to be sufficient. Training uses AdamW~\cite{loshchilov2019decoupledweightdecayregularization} with a 2e-5 learning rate for 3 epochs, optimizing a standard cross-entropy loss.

Our primary benchmark is CinePile~\cite{rawal2024cinepilelongvideoquestion}, where we report average and category-wise accuracy. For generalization, we perform zero-shot evaluation on ActivityNet-QA~\cite{yu2019activitynetqadatasetunderstandingcomplex}. Since ActivityNet-QA is open-ended, we adapt it to a multiple-choice format by generating four distractors per question using GPT-4o~\cite{gao2023palprogramaidedlanguagemodels}, ensuring consistency with our training setup. Baselines include zero-shot InternVL2-2B, larger models (mPLUG-Owl3, LongVA), and our model fine-tuned without CoT.

\subsection{Implementation Details}

InternVL2-2B uses a ViT-based visual encoder~\cite{dosovitskiy2021imageworth16x16words} and a LLaMA-style decoder~\cite{touvron2023llamaopenefficientfoundation}. Due to backpropagation constraints~\cite{gh-ft-issue}, we use 3 frames per video during fine-tuning and 12 during inference. The pipeline is implemented in PyTorch~\cite{paszke2019pytorchimperativestylehighperformance} and Hugging Face Transformers~\cite{wolf2020huggingfacestransformersstateoftheartnatural}. Fine-tuning completes in under 2 hours on a 4xA100 node. We plan to release our code to support reproducibility.
\section{Experiments and Results}
\label{sec:exp_results}
\begin{figure}[h]
    \centering
    \includegraphics[width=1.0\linewidth]{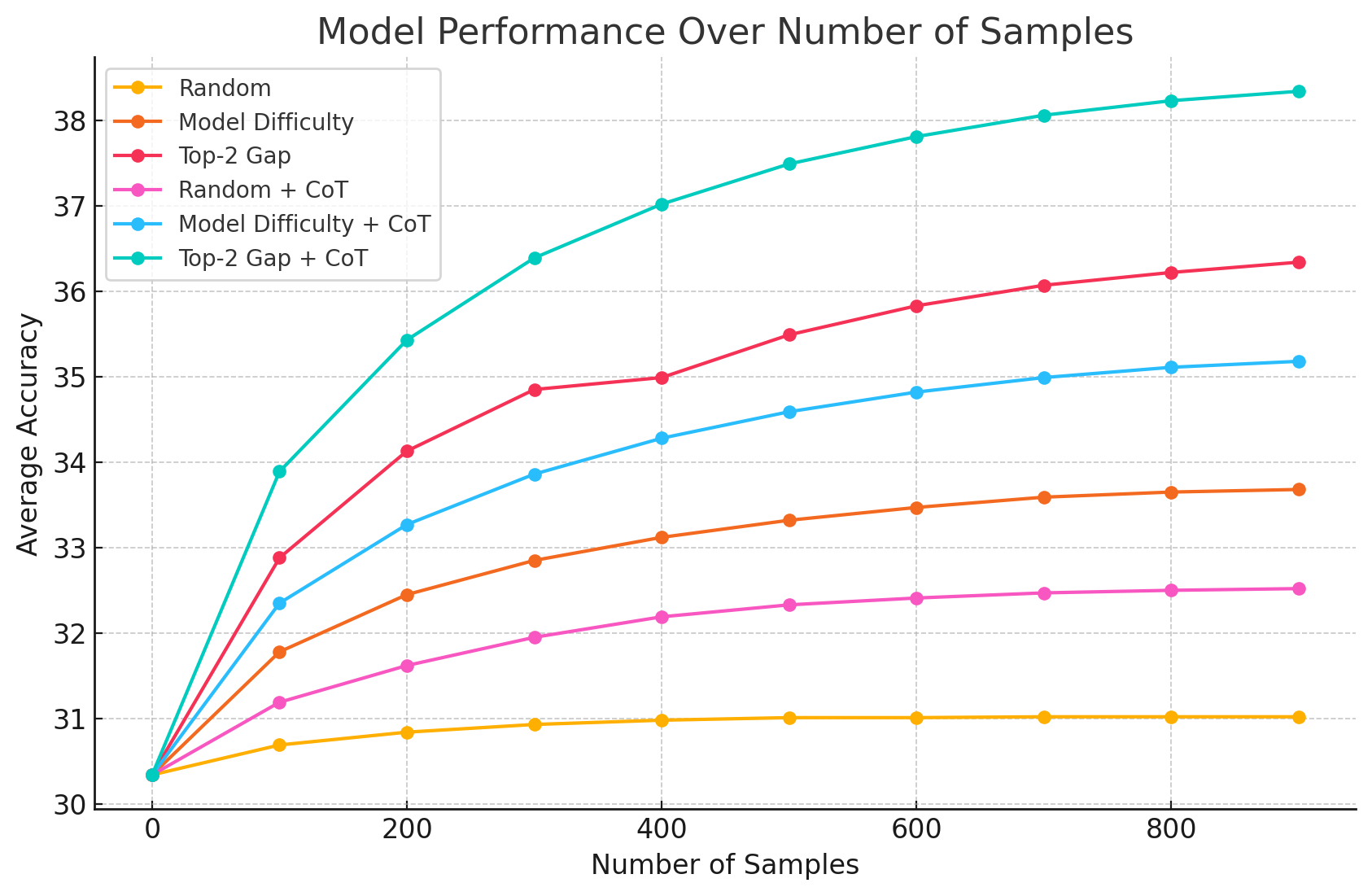}
    \caption{Average accuracy on CinePile dataset of InternVL2 2B models based on number of samples used for fine tuning.}
    \label{fig:linegraph1}
\end{figure}
We conduct experiments to evaluate our methodology for enhancing VLMs using targeted synthetic data. Our key research questions are:
\begin{itemize}
\item (Q1) How effective are different hard sample selection strategies compared to random sampling for fine-tuning efficiency?
\item (Q2) What is the impact of incorporating synthetically generated Chain-of-Thought (CoT) rationales, and how does the rationale integration format affect performance?
\item (Q3) Can our approach significantly narrow the performance gap between the compact InternVL2-2B model and larger VLMs on the challenging CinePile benchmark?
\item (Q4) Do the reasoning improvements learned on CinePile generalize to an unseen dataset (ActivityNet-QA) when evaluated in a zero-shot setting?
\end{itemize}

\begin{table*}[t]
\centering
\begin{tabular}{l|c|c|c|c|c|c|c}
\hline
Model & \#Params & Avg & CRD & NPA & STA & TEMP & TH \\
\hline
\hline 
InternVL2 2B & 2B & 30.34 & 31.91 & 33.26 & 30.35 & 23.26 & 31.58 \\
InternVL2 4B & 4B & 39.89 & 42.99 & 47.73 & 36.23 & 32.99 & 41.58 \\
mPLUG-Owl3 & 8B & 38.27 & 40.91 & 45.71 & 33.86 & 33.09 & 46.20 \\
InternVL2 8B & 8B & 32.28 & 35.25 & 40.39 & 28.46 & 24.71 & 38.42 \\
LongVA 7B & 7B & 41.04 & 43.28 & 51.84 & 38.45 & 33.58 & 38.42 \\
\hline 
InternVL2 2B (Random) & 2B & 31.02 & 31.16 & 31.96 & 31.62 & 26.06 & 40.21 \\
InternVL2 2B (Difficulty) & 2B & 33.68 & 34.45 & 38.15 & 33.10 & 27.74 & 40.48 \\
InternVL2 2B (Top-2 Gap) & 2B & 36.34 & 37.73 & 44.35 & 34.58 & 29.43 & 40.74 \\
\hline 
InternVL2 2B (Random + CoT) & 2B & 32.52 & 33.10 & 33.46 & 33.12 & 27.56 & 41.71 \\
InternVL2 2B (Difficulty + CoT) & 2B & 35.18 & 35.95 & 39.65 & 34.60 & 29.24 & 41.98 \\
\textit{InternVL2 2B (Top-2 Gap + CoT, 26B teacher)} & 2B & 37.41 & 38.45 & 47.17 & 35.62 & 30.12 & 42.21 \\
\textbf{InternVL2 2B (Top-2 Gap + CoT)} & \textbf{2B} & \textbf{38.34} & \textbf{39.73} & \textbf{46.50} & \textbf{36.58} & \textbf{30.85} & \textbf{42.74} \\
\hline
\end{tabular}
\caption{Performance Comparison of Vision-Language Models. We also compare synthetic CoT generated from a 26B teacher (italicized row) vs. 4B (final row). The 4B teacher outperforms, supporting its use as an efficient rationale generator.}
\label{tab:model_performance}
\end{table*}

\subsection {Experimental Setup}
\begin{itemize}
\item \textbf{Datasets}:
\begin {itemize}
    \item Cinepile \cite{rawal2024cinepilelongvideoquestion}: Our primary dataset for fine-tuning and evaluation. It comprises ~305,000 multiple-choice question-answer pairs derived from over 9,000 movie clips (avg. length ~160s). We utilize the official training split for fine-tuning sample selection and the official test split for evaluation. Its focus on long-form video and reasoning makes it a suitable testbed.
    \item ActivityNet-Qa \cite{yu2019activitynetqadatasetunderstandingcomplex}: Used for zero-shot generalization testing. It contains ~58k open-ended QA pairs on ~5.8k web videos (avg. length ~180s). We evaluate on the standard validation split. To align with our multiple-choice fine-tuning paradigm, we synthetically generated multiple-choice options for the ActivityNet-QA validation set using GPT-4o.
\end {itemize}

\item \textbf{Evaluation Metrics}: 
\begin {itemize}
    \item Cinepile: We report Average Accuracy across all questions and breakdown accuracy by the provided question types categories: \textbf{TEMP} (Temporal), \textbf{CRD} (Character and Relationship Dynamics), \textbf{NPA} (Narrative and Plot Analysis), \textbf{STA} (Setting and Technical Analysis), and \textbf{TH} (Thematic Exploration).
    \item ActivityNet-QA: We report standard multiple-choice Accuracy.
\end{itemize}
\item \textbf{Baselines and Compared Models}:
\begin{itemize}
    \item Base Model: InternVL2-2B (zero-shot) \cite{chen2024internvlscalingvisionfoundation}.
    \item Fine-tuned (No CoT): InternVL2-2B fine-tuned on N=900 samples selected via: (i) Random, (ii) Model-Estimated Difficulty (denoted Difficulty), (iii) Top-2 Confidence Gap (denoted Top-2 Gap).
    \item Fine-tuned (+Synth CoT): The same three selection strategies, but fine-tuned with synthetic CoT rationales generated by InternVL2-4B (denoted Random+CoT, Difficulty+CoT, Top-2 Gap+CoT). Rationales are integrated using the Answer-then-Rationale format determined in Section 3.4.
    \item Larger Models (for reference): InternVL2-4B, mPLUG-Owl3-8B \cite{ye2023mplugowl2revolutionizingmultimodallarge}, InternVL2-8B, LongVA-7B.
\end{itemize}
\end{itemize}

\subsection {Sample Efficiency Analysis}

To analyze data efficiency, we varied the number of fine-tuning samples (N) from 0 to 900 and tracked the Average Accuracy on the CinePile test set. Figure 2  plots the accuracy curves for Random selection, Top-2 Gap selection (without CoT), and Top-2 Gap selection with synthetic CoT. As discussed in section 3, we see the gains plateau at around N=700, and thus, we chose N=900 samples for fine tuning the models in the end.

The results clearly demonstrate the benefits of our approach. Firstly, targeted selection significantly improves sample efficiency. The Top-2 Gap strategy achieves higher accuracy with far fewer samples compared to Random selection. For instance, Top-2 Gap using only 300 samples (34.85\% Acc) surpasses the performance of Random selection using 900 samples (31.02\% Acc). These results highlight that CoT quality directly influences student performance. When the model is trained on unrelated CoT text, its accuracy drops below the base model, indicating that noisy supervision can be harmful. This supports our claim that carefully generated, instruction-following CoT rationales are crucial for effective transfer.
This confirms that focusing fine-tuning on points of model uncertainty accelerates learning. Secondly, incorporating synthetic CoT rationales provides a substantial additional boost in data efficiency. The Top-2 Gap+CoT curve consistently lies above the Top-2 Gap curve. Adding synthetic CoT rationales to 600 samples (37.81\% Acc) achieves performance comparable to using 900 samples without CoT (36.34\% Acc). This indicates that the synthetic reasoning signal effectively acts as potent data augmentation, allowing the model to extract more value from each selected sample. Notably, performance gains begin to saturate around N=700 samples for the targeted strategies, justifying our choice of N=900 for main experiments as an efficient operating point. We additionally ran a 1200-sample experiment and found performance gains began to plateau beyond 900 samples, supporting our claim that 900 is a sufficient budget. This is consistent with the flattening curve in Figure~\ref{fig:linegraph1}.

To better understand the impact of teacher model size on synthetic CoT quality, we compared the rationales generated by InternVL2-8B and InternVL2-26B. As shown in Table~\ref{tab:cot_complexity}, the larger teacher produces significantly longer CoTs with more reasoning-heavy tokens, which can overwhelm smaller students. These findings support our ``capacity mismatch'' hypothesis---that extremely verbose CoTs may hinder learning in low-capacity models like InternVL2-2B.

\begin{table}[t]
\centering
\begin{tabular}{l|c|c}
Teacher Model & Avg. CoT Length & \% Reasoning Tokens \\
\hline
InternVL2-8B & 25 tokens & 18\% \\
InternVL2-26B & 43 tokens & 27\% \\
\end{tabular}
\caption{Comparison of CoT complexity across teacher models. InternVL2-26B generates longer, denser rationales, which may introduce a mismatch when distilling into small students.}
\label{tab:cot_complexity}
\end{table}

\paragraph{Impact of CoT Quality.}
To evaluate how the quality of synthetic chain-of-thought (CoT) affects model performance, we conducted an ablation where the CoT rationales were replaced with randomly generated, unrelated text of similar length. This allows us to isolate the effect of structured reasoning. Results on CinePile are shown below:

\begin{table}[h]
\centering
\begin{tabular}{l|c}
Model Variant & Accuracy (\%) \\
\hline
InternVL2-2B (base) & 30.34 \\
+ Top-2 Gap + CoT (real) & 38.34 \\
+ Top-2 Gap + Random CoT & 28.00 \\
\end{tabular}
\caption{Ablation study showing the effect of CoT quality. Poor-quality (random) CoT hurts performance, while meaningful CoT improves accuracy.}
\label{tab:cot_quality}
\end{table}

These results validate that the observed improvements stem from the reasoning content in CoT, rather than merely increased sequence length or noise injection. In fact, unrelated rationales actively degrade model performance below baseline.

\subsection {Main Results on CinePile}

Table~\ref{tab:model_performance} presents the performance of different models and fine-tuning strategies on the CinePile test set.

The key observations from our main results (Table~\ref{tab:model_performance}) are:

\begin{itemize}
\item \textbf{Hard Sampling Effectiveness:} All fine-tuning approaches outperform the zero-shot baseline (30.34\%). Among the selection strategies without CoT, Top-2 Gap (36.34\%) significantly outperforms both Random (31.02\%, +5.32 points) and Difficulty (33.68\%, +2.66 points). This strongly supports selecting samples based on model uncertainty as the most effective strategy.

\item \textbf{Synthetic CoT Impact:} Adding synthetic CoT rationales consistently improves performance. The largest gain is observed for the best selection method: Top-2 Gap+CoT achieves 38.34\%, a +2.0 point improvement over Top-2 Gap without CoT. The boost is particularly pronounced in reasoning-intensive categories like TEMP and CRD, demonstrating that the synthetic rationales successfully impart reasoning skills.

\item \textbf{Closing the Gap:} Our best model, InternVL2-2B (Top-2 Gap+CoT), reaches 38.34\% average accuracy. This significantly closes the gap to the larger InternVL2-4B baseline (39.89\%), reducing the difference from 9.55 points to just 1.55 points. Notably, it surpasses the accuracy of the mPLUG-Owl3-8B model (38.27\%) and the InternVL2-8B model (32.28\%). While still below the specialized LongVA-7B (41.04\%), this highlights the power and practicality of our methodology.
\end{itemize}

\subsection{Generalization to MLVU}
\label{sec:mlvu}

To provide the strongest test of generalization, we evaluated our model on a challenging held-out subset of the MLVU benchmark, which features long-form videos and compositional question types. The results, shown in Table~\ref{tab:mlvu}, demonstrate a remarkable feat of knowledge distillation. Our fine-tuned 2B model achieves an impressive 50.5\% accuracy, not only representing a massive 5.5-point leap over its base model (45.0\%) but also achieving near-parity with its 4B teacher (51.0\%). By closing the performance gap to a razor-thin 0.5\% margin against a model twice its size, this striking result validates that our efficient, targeted fine-tuning strategy successfully distills complex reasoning capabilities into a far more compact architecture.

\begin{table}[h]
\centering
\caption{Accuracy on a subset of the MLVU benchmark.}
\begin{tabular}{l|c}
\hline
Model & Accuracy (\%) \\
\hline
InternVL2-2B (base) & 45.0 \\
InternVL2-2B (Top-2 Gap + CoT) & \textbf{50.5} \\
InternVL2-4B (base) & 51.0 \\
\hline
\end{tabular}
\label{tab:mlvu}
\end{table}

\subsection {Generalization to ActivityNet-QA (Zero-Shot Transfer)}
To further confirm the transferability of these gains, we tested our model zero-shot on a synthetically adapted ActivityNet-QA multiple-choice benchmark. The baseline model achieves 40.20\% accuracy. Fine-tuning on CinePile with hard sample selection alone provides a modest gain (41.46\%). However, adding our synthetic CoT rationales results in a more substantial improvement, reaching 43.19\% accuracy (+2.99 points over baseline, +1.73 points over fine-tuning without CoT). This demonstrates that the enhanced reasoning capabilities generalize robustly to a different dataset and task format, confirming our approach imparts transferable skills.

\paragraph{Justification for Synthetic Evaluation on ActivityNet-QA} 
To align the open-ended ActivityNet-QA with our multiple-choice setup, we used GPT-4o to generate four semantically plausible and contextually relevant distractors for each question. This process is detailed with an example in the supplementary material to illustrate the quality of our evaluation protocol.
\section{Discussion and Limitations}
\label{sec:discussions_limitations}

Our research shows that strategically generated and selected synthetic data enhances reasoning in compact Vision-Language Models (VLMs). By fine-tuning InternVL2-2B on a targeted subset ($\sim$0.3\%) of CinePile data with synthetic Chain-of-Thought (CoT) rationales, we achieved performance approaching a 2x larger model and surpassed other large baselines, demonstrating a highly efficient path to model improvement.

\begin{itemize}
    \item \textbf{Synergistic Efficiency of Selection and Synthetic CoT:} A key finding is the synergy between hard sample selection (Top-2 Gap) and synthetic CoT supervision. Selection significantly improves data efficiency (Fig.~\ref{fig:linegraph1}), while synthetic rationales amplify learning on high-uncertainty samples where explicit reasoning is most valuable (Table~\ref{tab:model_performance}). These rationales help compact models disentangle temporal dependencies, a key challenge for small architectures.

    \item \textbf{Moderate Teachers Enable Efficient Synthetic Reasoning:} Our finding that a 4B teacher outperforms a 26B teacher for this task underscores the value of intermediate teacher models. This supports the hypothesis that overly complex rationales from large teachers can overwhelm small models, and that moderate-capacity teachers provide a better trade-off between quality and distillability \cite{li2025smallmodelsstrugglelearn}, significantly reducing compute requirements for data generation.

    \item \textbf{Sensitivity to Synthetic Data Format:} The success of our \\ \verb|Answer-then-Rationale| format over the standard \verb|Rationale-then-Answer| (which caused an "Option E bias") reveals that compact models are highly sensitive to the structure of synthetic supervision. We hypothesize this stems from the cognitive burden of pre-answer reasoning on low-capacity models, token decay in long outputs, and more effective optimization when supervision focuses early on the answer. This suggests that \emph{post-hoc rationalization is not only easier to learn, but more effective}, motivating \textbf{format design tailored to model capacity}.
\end{itemize}

In summary, our results motivate a reevaluation of standard CoT prompting in low-resource settings. Simply reordering the rationale can yield significant gains, offering a practical path for boosting compact VLMs.

\subsection*{Limitations}

\begin{itemize}
    \item \textbf{Teacher Quality and Over-Reliance:} The effectiveness of synthetic CoT is bounded by the teacher's quality. Errors or hallucinations from the 4B teacher can propagate through the synthetic data \cite{perez2022discoveringlanguagemodelbehaviors}, underlining the need for careful auditing and filtering for rationale faithfulness \cite{lanham2023measuringfaithfulnesschainofthoughtreasoning}, especially in high-stakes settings.

    \item \textbf{Robustness and Generalizability Scope:} Our sample selection relies on model confidence, which is not always well-calibrated \cite{guo2017calibrationmodernneuralnetworks}. While we show generalization to ActivityNet-QA and MLVU, further validation across different architectures is needed. Additionally, our ActivityNet-QA evaluation uses GPT-generated options, a synthetic layer that could influence outcomes.

    \item \textbf{Edge Deployment and Applicability:} Our experiments use InternVL2-2B, as few compact VLMs handle temporal reasoning (e.g., Qwen-VL-Tiny is image-focused). Larger teachers (>4B) are also infeasible for on-device generation. Our method serves as a template for future compact VLMs, and extending it to new architectures remains a key future direction.
\end{itemize}
\section{Conclusion}
\label{sec:conclusion}

We proposed an effective pipeline to enhance compact VLMs using synthetic Chain-of-Thought (CoT) rationales and targeted fine-tuning. By selecting just $\sim$900 high-uncertainty samples and generating concise rationales from a moderate (4B) teacher, we significantly boosted InternVL2-2B's performance on the CinePile benchmark, outperforming larger models. Our method demonstrates strong zero-shot generalization to ActivityNet-QA and near-parity with its 4B teacher on the MLVU benchmark. Ablations highlight the value of capacity-aware distillation, showing a 4B teacher can outperform a 26B model in this setting. By achieving performance rivaling 4$\times$ larger models with minimal data and compute, our work enables the practical deployment of reasoning-rich VLMs on edge devices. We will release our code upon publication to encourage broader exploration of targeted synthetic augmentation for small multimodal models.

\subsubsection*{Disclosure of Interests}
The authors have no competing interests to declare that are relevant to the content of this article.
\bibliographystyle{splncs04}
\bibliography{main}
\end{document}